# Fostc3net：A Lightweight YOLOv5 Based On the Network Structure Optimization

Danqing Ma[1*], Shaojie Li[2], Bo Dang[3], Hengyi Zang[4], Xinqi Dong[5]

[1]Computer Science, Stevens Institute of Technology, Hoboken NJ, U.S,
[2]Computer Technology, Huacong Qingjiao Information Technology Co., Ltd, Beijing, China
[3]Computer Science, San Francisco Bay University, Fremont CA, U.S
[4]Physics and Mathematics, Universidad Tecnológico Universitam, Beijing, China.
[5]Computer Science, University of Maine at Presque Isle, Presque Isle ME, U.S

[*]The corresponding author email: 3530761316qq@gmail.com

**Abstract.** Transmission line detection technology is crucial for automatic monitoring and ensuring the safety of electrical facilities. The YOLOv5 series is currently one of the most advanced and widely used methods for object detection. However, it faces inherent challenges, such as high computational load on devices and insufficient detection accuracy. To address these concerns, this paper presents an enhanced lightweight YOLOv5 technique customized for mobile devices, specifically intended for identifying objects associated with transmission lines. The C3Ghost module is integrated into the convolutional network of YOLOv5 to reduce floating point operations per second (FLOPs) in the feature channel fusion process and improve feature expression performance. In addition, a FasterNet module is introduced to replace the c3 module in the YOLOv5 Backbone. The FasterNet module uses Partial Convolutions to process only a portion of the input channels, improving feature extraction efficiency and reducing computational overhead. To address the imbalance between simple and challenging samples in the dataset and the diversity of aspect ratios of bounding boxes, the wIoU v3 LOSS is adopted as the loss function. To validate the performance of the proposed approach, Experiments are conducted on a custom dataset of transmission line poles. The results show that the proposed model achieves a 1% increase in detection accuracy, a 13% reduction in FLOPs, and a 26% decrease in model parameters compared to the existing YOLOv5.In the ablation experiment, it was also discovered that while the Fastnet module and the CSghost module improved the precision of the original YOLOv5 baseline model, they caused a decrease in the mAP@.5-.95 metric. However, the improvement of the wIoUv3 loss function significantly mitigated the decline of the mAP@.5-.95 metric.

**Keywords:** Improved YOLOv5 , FasterNet , GhostNet, Lightweighting.

## 1. Introduction

The expansion of the power system has heightened security concerns regarding high-voltage, long-distance overhead transmission lines. Damage caused by birds, kites, and balloons can pose significant risks to the safety and stability of these lines[1]. Therefore, intelligent monitoring of the areas surrounding transmission lines is crucial. Timely and precise detection of potential threats is crucial to ensure the uninterrupted and secure functioning of these transmission lines.

Deep learning has become one of the current focuses of research in many industries [2]. In particular, the emergence of convolutional neural networks has given computer vision new vitality in different fields [3]. In recent years, with the continuous development of the field of computer vision, computer vision based on the attention mechanism of the transfermer architecture is also showing its prominence in various fields such as medical treatment [4] and autonomous driving [5].

In the early stages, two-step region-based convolutional neural network target detection methods such as FasterR-CNN were used. However, due to the large monitoring area of transmission lines and the reality that most abnormal targets are small, the slow detection speed of the two-step methods could not meet the high response requirements of daily monitoring systems. Subsequently, regression-based target recognition methods such as SSD and Yolo have significantly improved detection efficiency by leveraging their end-to-end detection and recognition features[6]. YoloV5 algorithm, in particular, has become a mature and mainstream detection method.

With the gradual application of the yolov5 algorithm, lightweighting has become one of the focuses of development and optimization. The lightweight algorithm can have faster response speed, lower resource requirements and lower energy consumption, which is undoubtedly important for transmission line foreign object detection equipment that needs to be deployed on a large scale. Huang et al. developed a method that integrates the parameter-free attention module, SimAM, into the YOLOv5 network to enhance its feature extraction capability.This enables the detection network to more efficiently identify a variety of foreign objects on the transmission line [7]. Lu et al. reconstructed the C3 module of the YOLOv5 model using GhostNetV2 for the specific application of detecting foreign objects on transmission lines[8]. Peng et al. adopted a new bounding box loss function, Focal-CIoU, to improve the precision of lightweight models.This new loss function is particularly suitable for the detection of electrical lines where there is a large gap between foreign objects and surroundings.[9]. The improvements made by the above researchers have made outstanding contributions to the detection accuracy of transmission lines and the lightweight of equipment.

Despite these advancements, some issues such as inadequate structural modifications of the grid still persist. These issues result in insufficient model lightweighting and substantial computational burdens. To address those problems, this study introduces an enhanced YOLOv5 network structure named Fost-c3net, designed specifically for lightweight purposes.In order to enhance the feature extraction process of images in the network's foundation, the C3 module in the original YOLOv5 Backbone is replaced with the FasterNet. The GhostBottleneck replaces the Bottleneck in the original YOLOv5 Neck network's C3 module with a new C3Ghost module, resulting in enhanced running speed due to model compression and lower computational demands. Them,the loss function is Replaced with WIoU (Weighted Intersection over Union) in place of the original CIoU (Complete Intersection over Union) loss function addressed the issue of imbalance between easy and hard samples in the dataset. This architectural change not only enhanced the accuracy but also significantly reduced the parameters and FLOPs within the structure, demonstrating the superior performance of Fost-c3net in detecting defects in power transmission lines.

## 2. Methodology

This section will introduce three network structure changes based on lightweight yolov5.To improve the feature extraction process of images in the network backbone, the original C3 module in YOLOv5 has been replaced with a FasterNet module. The architecture of FasterNet[10], as depicted in Figure 1, adopts a novel form of convolution. It integrates Partial Convolution with two pointwise convolutions, utilizing an inverse residual connection to effectively address the issue of vanishing gradients. The PConv component extracts spatial features from only a subset of input channels while preserving the integrity of the remaining channels. By processing only a fraction of the channels, this approach addresses the problem of low floating point operations per second (FLOPS) caused by frequent memory accesses by the operators.

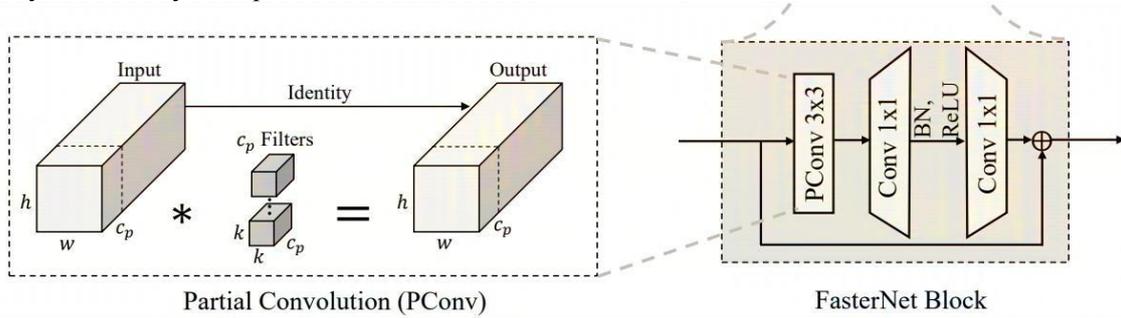

**Figure 1**. FasterNet

The original YOLOv5 Neck network's Bottleneck is substituted with the innovative C3Ghost module from GhostNet [11]. The core design principle of GhostNet is to generate the majority of feature maps with fewer convolutional operations, thereby reducing computational cost and model parameters. Initially, a set of feature maps is generated through standard convolutional operations. Subsequently, these feature maps are subject to a cheap operation to produce redundant feature maps, generating additional Ghost feature maps. This step utilizes depthwise (DW) convolutions derived from simple linear transformations. Finally, the feature maps generated by the standard convolution are concatenated with those generated through the cheap operation. This approach enables GhostNet to acquire a broader range of feature insights with lower computational demand, resulting in model lightweighting and improved performance.

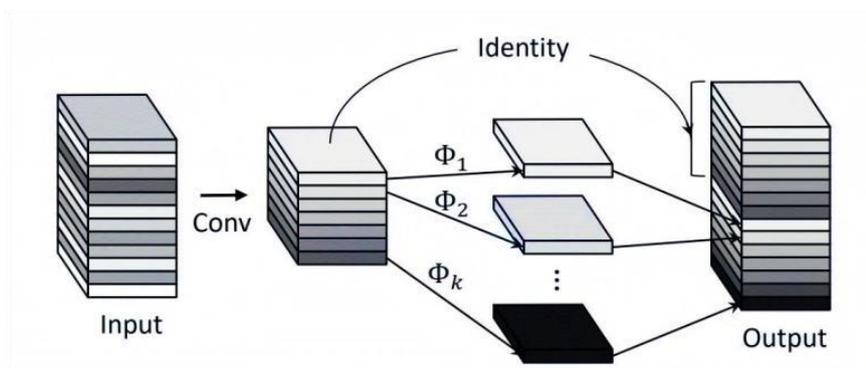

**Figure 2.** GhostNet

The third improvement involves replacing the CIoU loss function with the WIoU loss function to address the imbalance between easy and difficult samples in the dataset [12]. It is important to note that in the context of bounding box regression, if the aspect ratio of the predicted and actual boxes is linear, the relative ratio penalty in the CIoU Loss becomes nullified, making it ineffective. Consequently, the WIoU loss takes into account both data quality and diversity in aspect ratios, providing a more balanced and comprehensive approach. This methodology significantly improves the model's robustness and accuracy when dealing with varying data conditions, optimizing overall performance.The complete Fostc3net structure is shown in Figure 3.

## 3. Experimental results

We collect a dataset for foreign object detection on transmission lines, which includes various categories of targets such as balloons, kites, bird nests, and trash. There are 731 pictures in total, including 427 instances of bird nests, 231 kites, 192 balloons and 231 trash.To train the model for detecting transmission lines, the constructed dataset is divided into a training set, a validation set,and a test set at a ratio of 7:2:1.

Performance metrics include Precision, which is the proportion of true positives among the predicted positives, and Recall, which is the proportion of true positives among all actual positives.The IoU threshold used is 0.5. They also include mAP@.5 and mAP@.5-.95, which represent the mean Average Precision at various Intersections over Union thresholds, reflecting the model's detection capabilities under diverse conditions. Model complexity indicators include the number of parameters, Giga Floating Point Operations(GFLPOs). High Precision and Recall indicate great recognition capabilities of the model, while high mAP@.5 and mAP@.5-.95 values show robust performance across a spectrum of IoU thresholds. The final performance is shown in Table 1, The training results of Fostc3net are shown in Figure 4. The actual detection results of Fostc3net on the mobile terminal are shown in Figure 5.

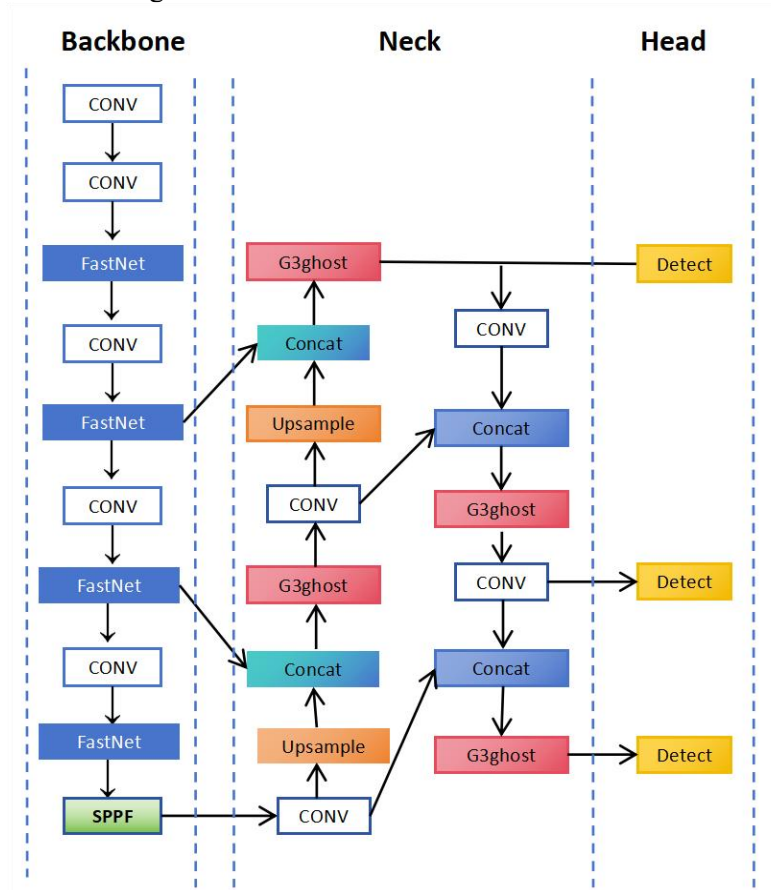

**Figure 3**. The structure of Fostc3net

**Table 4.** Model training time results

| Model | Precision | Recall | mAP@.5 | mAP@.5-.95 | Parameters(M) | GFLOPs |
|---|---|---|---|---|---|---|
| YOLOv5 | 0.968 | 0.977 | 0.987 | 0.714 | 7.102 | 14.82 |
| Fostc3net | 0.979 | 0.981 | 0.988 | 0.711 | 5.212 | 12.88 |

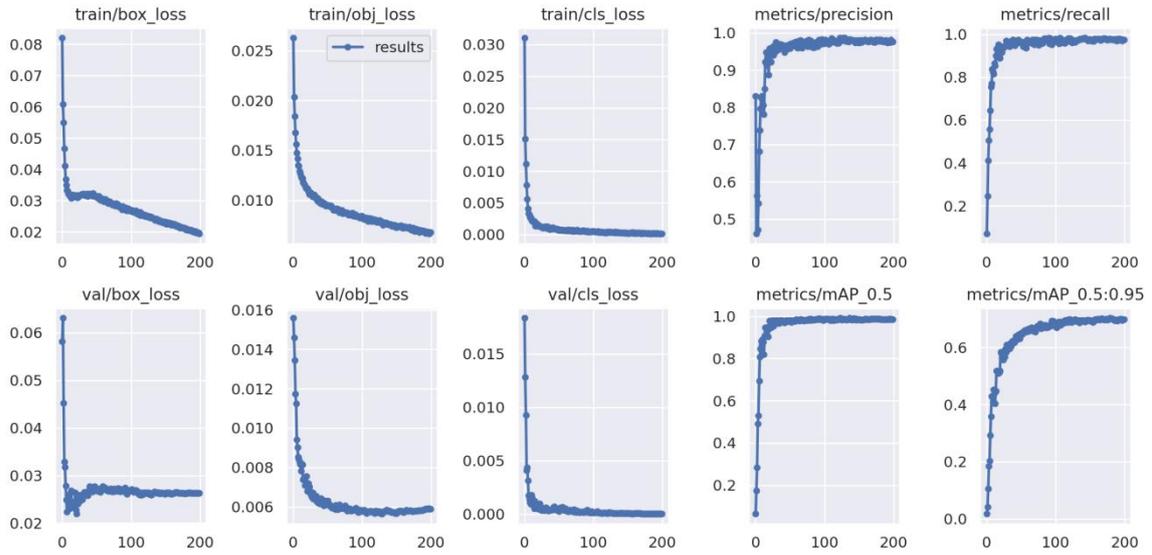

**Figure 4**. The training process of Fostc3net

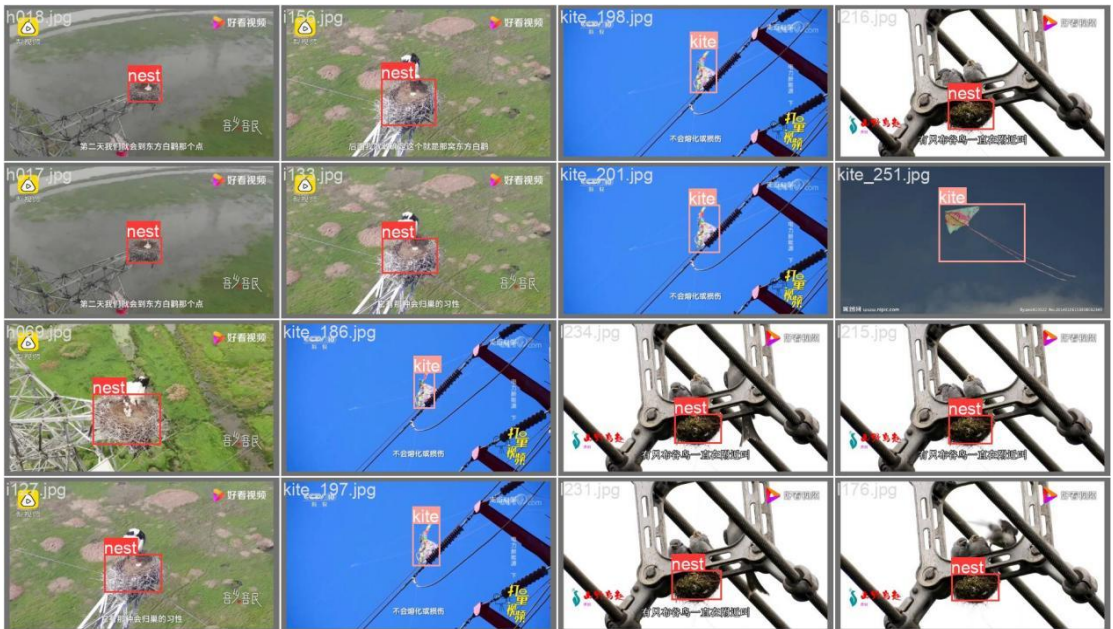

**Figure 5.** The actual effect after Fostc3net training

Table 4 presents a performance comparison between our model and the benchmark model across various metrics. The results show that, compared to the YOLOv5 model, our model achieves approximately a 1.14% increase in Precision, a 0.41% increase in Recall, and a 0.10% increase in mAP@.5, but a decrease of about 0.42% in mAP@.5-.95. These findings suggest that our model performs well under the default IoU thresholds, but experiences a slight decline under stricter evaluation criteria.However, since our goal is to maintain a lightweight approach, the performance improvement may be attributed to the random selection of the training set and the test machine. Overall, the improved model does not result in a decrease in performance. However, it is important to combine the specific evaluation with lightweight indicators.It is noteworthy that the number of parameters in our model is reduced by approximately 26.61%, and GFLOPs by about 13.09%. This demonstrates that the Fostc3net model is more lightweight.

Based on the results, we draw a preliminary conclusion that the improved model has been significantly improved. However, it is crucial to determine the impact of the network structure based on the slight improvements on the performance of the model. Therefore, we conducted ablation experiments, as shown in Figure 6, to analyze the effect of each part of the improved network compared to the final complete network structure. The ablation experiment used the 'all' category of metrics as an example.

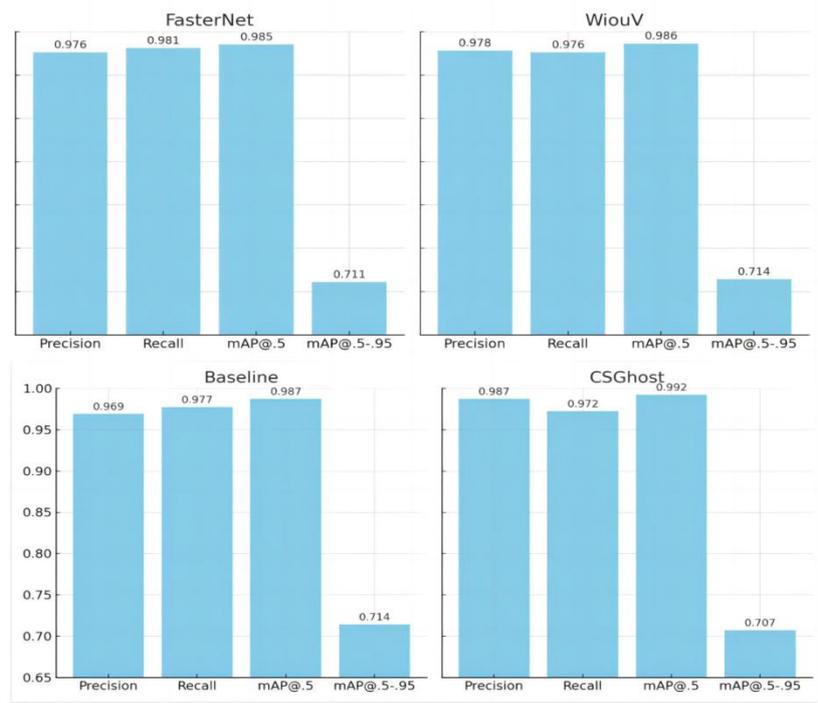

**Figure 6. Ablation Experiments Results for Each Part**

In ablation experiments, the CSGhost structure performs best in terms of accuracy, while the baseline model has the best recall, and WiouV3 shows the smallest performance improvement. This demonstrates that the performance of CSGhost, based on neck modification, exceeds that of Fastnet on the backbone network. We also find that switching to a neural network model with a different structure causes a slight decline in the mAP@.5-.95 indicator, and switching to a different loss function can largely alleviate this decline. This indicates that the main changes in WiouV3 are concentrated in the loss function, which typically only has an indirect impact on model performance. This impact is reflected in a broader and more accurate representation of localization precision compared to traditional distance-based loss functions. To verify this finding, we conduct further ablation experiments, retaining the previously improved structure and controlling for loss function variables. The results are as expected. Compared with Fostc3net, the accuracy of the model without the WiouV3 loss function does not improve, while mAP@.5-.95 decreases by 2 percentage points. This shows that the addition of the loss function does improve the mAP@.5-.95 of the model without affecting the accuracy.

## 4. Conclusion

This article proposes an improved yolov5 model Fostc3net based on lightweight purposes. It can significantly reduce the required parameters and computing load without reducing detection performance, and can run well on lightweight devices. The results indicate that the designed model achieves a 1% increase in detection accuracy, a 13% reduction in FLOPs, and a 26% improvement in model parameters compared to the existing YOLOv5.In the ablation experiment, it was found that the addition of the structures of Fastnet and CSghost improved the precision of yolov5, but the mAP@.5-.95 decreased. The addition of the Wiou loss function improves the reduced mAP@.5-.95 index. Although Fostc3net has demonstrated superior performance, more types of data sets for detecting small targets are still needed for more comprehensive verification. The introduction of wiouv3 loss function only alleviates or eliminates the decrease in mAP@.5-.95 index caused by the introduction of lightweight modules, and does not achieve substantial improvement. Therefore, in the future lightweight transformation, in-depth research on network structure transformation will continue to pursue greater performance improvement. Future research can also consider using other types of lightweight methods, such as model distillation, which will also help further compress model size and improve performance.

**References**
[1] Bao W, Du X, Wang N, Yuan M, Yang X 2022 A Defect Detection Method Based on BC-YOLO for Transmission Line Components in UAV Remote Sensing Images Remote Sensing 14 5176.
[2] Liu C, Pang Z, Ni G, Mu R, Shen X, Gao W, Miao S 2023 A comprehensive methodology for assessing river ecological health based on subject matter knowledge and an artificial neural network Ecological Informatics 77 102199.
[3] Liu C, Li H, Xu J, Gao W, Shen X, Miao S 2023 Applying Convolutional Neural Network to Predict Soil Erosion: A Case Study of Coastal Areas International Journal of Environmental Research and Public Health 20(3) 2513.
[4] Dai W, Mou C, Wu J, Ye X 2023 Diabetic Retinopathy Detection with Enhanced Vision Transformers: The Twins-PCPVT Solution In 2023 IEEE 3rd International Conference on Electronic Technology, Communication and Information (ICETCI) (pp. 403-407) IEEE.
[5] Mou C, Dai W, Ye X, Wu J 2023 Research On Method Of User Preference Analysis Based on Entity Similarity and Semantic Assessment In 2023 8th International Conference on Signal and Image Processing (ICSIP) (pp. 1029-1033) IEEE.
[6] Zhang X, Zhang L, Li D 2019 Transmission line abnormal target detection based on machine learning yolo v3 In 2019 International Conference on Advanced Mechatronic Systems (ICAMechS) (pp. 344-348) IEEE.
[7] Huang H, Lan G, Wei J, Zhong Z, Xu Z, Li D, Zou F 2023 TLI-YOLOv5: A Lightweight Object Detection Framework for Transmission Line Inspection by Unmanned Aerial Vehicle Electronics 12(15) 3340.
[8] Lu L, Chen Z, Wang R, Liu L, Chi H 2023 Yolo-inspection: defect detection method for power transmission lines based on enhanced YOLOv5s Journal of Real-Time Image Processing 20(5) 104.
[9] Peng H, Liang M, Yuan C, Ma Y 2023 EDF-YOLOv5: An Improved Algorithm for Power Transmission Line Defect Detection Based on YOLOv5 Electronics 13(1) 148.
[10] Chen J, Kao S H, He H, Zhuo W, Wen S, Lee C H, Chan S H G 2023 Run, Don't Walk: Chasing Higher FLOPS for Faster Neural Networks In Proceedings of the IEEE/CVF Conference on Computer Vision and Pattern Recognition 12021-12031.
[11] Han K, Wang Y, Tian Q, Guo J, Xu C, Xu C 2020 Ghostnet: More features from cheap operations In Proceedings of the IEEE/CVF conference on computer vision and pattern recognition 1580-1589.
[12] Cho Y J 2021 Weighted Intersection over Union (wIoU): A New Evaluation Metric for Image Segmentation arXiv preprint arXiv:2107.09858.